\documentclass[final,1p,times,preprint,authoryear]{elsarticle}

\usepackage{subfig}
\usepackage{url}
\usepackage{pdflscape}
\usepackage[detect-weight=true, detect-family=true,binary-units=true]{siunitx}
\usepackage{microtype}
\usepackage{csquotes}
\usepackage{amsmath}

\usepackage{amssymb}

\usepackage{lineno}
\usepackage{booktabs}
\usepackage{multirow}

\usepackage[rgb]{xcolor}
\definecolor{codegreen}{rgb}{0,0.6,0}
\definecolor{codegray}{rgb}{0.5,0.5,0.5}
\definecolor{codepurple}{rgb}{0.58,0,0.82}

\definecolor{sklearnorange}{rgb}{0.97,0.6,0.22}
\definecolor{sklearnblue}{rgb}{0.19,0.60,0.81}

\usepackage{tikz}
\usetikzlibrary{positioning,shapes,arrows,intersections}
\tikzstyle{abstract}=[rectangle, draw=black, rounded corners, fill=sklearnblue, fill opacity=.7,
        align=center, anchor=north, text=black, text width=4.75cm]
\tikzstyle{extension}=[rectangle, draw=black, rounded corners, fill=sklearnorange, fill opacity=.7,
        align=center, anchor=north, text=black, text width=4.75cm]
\tikzstyle{line}=[-, very thick]
\usepackage{pgfplots}
\pgfplotsset{compat=newest}

\usepackage{listings}
\lstdefinestyle{mystyle}{
  commentstyle=\color{codegreen},
  keywordstyle=\color{magenta},
  numberstyle=\tiny\color{codegray},
  stringstyle=\color{codepurple},
  basicstyle=\ttfamily\footnotesize,
  breakatwhitespace=false,         
  breaklines=true,                 
  captionpos=b,                    
  keepspaces=true,                 
  numbers=left,                    
  numbersep=5pt,                  
  showspaces=false,                
  showstringspaces=false,
  showtabs=false,                  
  tabsize=2
}
\usepackage[breaklinks=true,colorlinks=true,linkcolor=blue,urlcolor=blue,citecolor=blue]{hyperref} 

\journal{Engineering Applications of Artificial Intelligence}

\begin{document}

\begin{frontmatter}



\title{PyRCN: A Toolbox for Exploration and Application of Reservoir Computing Networks}
\author[1]{Peter Steiner\corref{CorrespondingAuthor}\fnref{EqualContribution}}
\ead{peter.steiner@tu-dresden.de}
\author[2,3]{Azarakhsh Jalalvand\fnref{EqualContribution}}
\ead{azarakhsh.jalalvand@ugent.be}
\author[1]{Simon Stone}
\ead{simon.stone@tu-dresden.de}
\author[1]{Peter Birkholz}
\ead{peter.birkholz@tu-dresden.de}
\cortext[CorrespondingAuthor]{Corresponding author}
\fntext[EqualContribution]{Equal contribution.}

\address[1]{Institute for Acoustics and Speech Communication, Technische Universität Dresden, Dresden, Germany}
\address[2]{IDLab, Ghent University--imec, Ghent, Belgium}
\address[3]{Mechanical and Aerospace Engineering department, Princeton University, New Jersey, USA}

\begin{abstract}
    Reservoir Computing Networks (RCNs) belong to a group of machine learning techniques that project the input space non-linearly into a high-dimensional feature space, where the underlying task can be solved linearly. Popular variants of RCNs are capable of solving complex tasks equivalently to widely used deep neural networks, but with a substantially simpler training paradigm based on linear regression. In this paper, we show how to uniformly describe RCNs with small and clearly defined building blocks, and we introduce the Python toolbox PyRCN (Python Reservoir Computing Networks) for optimizing, training and analyzing RCNs on arbitrarily large datasets. The tool is based on widely-used scientific packages and complies with the scikit-learn interface specification. It provides a platform for educational and exploratory analyses of RCNs, as well as a framework to apply RCNs on complex tasks including sequence processing. With a small number of building blocks, the framework allows the implementation of numerous different RCN architectures. We provide code examples on how to set up RCNs for time series prediction and for sequence classification tasks. PyRCN is around ten times faster than reference toolboxes on a benchmark task while requiring substantially less boilerplate code.
\end{abstract}



\end{frontmatter}


\section{Introduction}
\label{sec:Introduction}

Reservoir Computing Networks (RCNs) summarize a variety of machine learning techniques that use random, non-linear projections of inputs into a high-dimensional feature space. This often greatly facilitates classification tasks, because classes that are not linearly separable in the original input space, may become linearly separable in the high-dimensional space. The simplest Reservoir Computing architecture is the Extreme Learning Machine (ELM) \citep{src:Huang-06}, which is similar to a conventional Feed-Forward neural network. The more common RCN architectures, however, are Echo State Networks (ESNs) \citep{src:Jaeger-01a} and Liquid State Machines (LSMs) \citep{src:Maass-02}, which are variants of Recurrent Neural Networks (RNNs). The basic idea of RCNs is rather simple: The input nodes are randomly connected to a single hidden layer, the so-called \enquote{reservoir}, consisting of non-linear neurons. Only connections from the hidden layer to the output are trained, typically using linear regression. In case of ELMs, the neurons in the hidden layer are not inter-connected and the basic ELM is thus closely related to Multilayer Perceptrons (MLPs). In case of ESNs, the hidden layer is a pool of randomly interconnected non-linear neurons. LSMs are essentially ESNs with spiking neuron models. Based on the number of citations of their landmark papers, RCNs are not as commonly used as other types of neural networks, such as MLPs, Convolutional Neural Networks (CNNs) or networks consisting of Long-Short-Term-Memory (LSTM) cells, but they have achieved results that rival those and more popular deep-learning architectures. For example, ESNs have been successfully used for speech \citep{src:Triefenbach-13}, image \citep{src:Jalalvand-18}, radar \citep{src:Jalalvand-19}, plasma control \citep{src:Jalalvand-21}, music \citep{src:Steiner-20b, src:Steiner-20a}, time-series prediction \citep{src:Trierweiler-Ribeiro-21,src:Rodrigues-Moreno-20}, electrical load and energy (consumption) forecasting \citep{src:Mansoor-21,src:da-Silva-21,src:Wang-18} and for robot control \citep{src:Salmen-05,src:Oubbati-05,src:Schrauwen-07,src:Antonelo-08,src:Antonelo-15}. ELMs have been successfully used for image classification \citep{src:Huang-06}, gesture recognition \citep{src:Katilmis-21}, pressure prediction \citep{src:Cocco-Mariani-19}, heartbeat classification \citep{src:Ding-14}, and for non-linear regression \citep{src:Tang-16}.

One important obstacle for RCNs on their way towards the mainstream may be the lack of a framework that allows a fair and straight-forward comparison of RCNs with other machine learning architectures. RCNs are neither part of any widely used Machine/Deep Learning toolbox for Python, such as scikit-learn \citep{src:Pedregosa-11}, PyTorch \citep{src:Paszke-19} or Tensorflow \citep{src:Abadi-16}, nor of Matlab. While specialized toolboxes have been developed in the past to optimize, train and evaluate RCNs, a side-by-side comparison of RCN models and traditional machine learning or deep-learning models was laborious. Interested users needed to change significant parts of their code or data structure, because many toolboxes neither provide a high-level interface nor utilize a common data structure. 

Table \ref{tab:OverviewRCNToolboxesLandscape} summarizes various RCN toolboxes and the here proposed PyRCN along with important aspects, such as download links, which kind of RCNs are implemented, the year of the last major update and whether the toolboxes provide a high-level interface. In general, most of the toolboxes are developed for only one specific type of RCN. In fact, OGER \citep{src:Verstraeten-12} was the only toolbox that implemented all important variants of RCNs. However, it is now out-of-date, especially because it was implemented in Python 2, which has reached its end of life in 2020. Only a few ESN toolboxes provide a high-level interface, which help the users to quickly try out ESNs. In fact, even if an ESN toolbox provides a high-level interface, it is often not compatible with the commonly used Machine/Deep Learning frameworks, such as scikit-learn, PyTorch or Tensorflow. The ELM toolboxes both provide a high-level interface that is compatible with scikit-learn. However, only the HP-ELM \citep{src:Akusok-15} is scalable towards large datasets and high-performance clusters, for which it provides specialized implementations. The Python-ELM \citep{src:Python-ELM} can only be used for classification tasks. Compared to the other presented toolboxes in Table \ref{tab:OverviewRCNToolboxesLandscape}, PyRCN is the only actively maintained toolbox that efficiently implements both ESNs and ELMs. While ReservoirPy by \citet{src:Trouvain-20} provides a high-level interface to its ESN implementation, it does not adhere to any established specification. Therefore, PyRCN is the only toolbox that provides an RCN implementation with a high-level interface that complies with the scikit-learn specification. Herbert Jaeger\footnote{\href{https://www.ai.rug.nl/minds/research/esnresearch/}{https://www.ai.rug.nl/minds/research/esnresearch/}}, Claudio Gallicchio\footnote{\href{https://sites.google.com/site/cgallicch/resources}{https://sites.google.com/site/cgallicch/resources}} and the IEEE Task Force on Reservoir Computing\footnote{\href{https://sites.google.com/view/reservoir-computing-tf/resources}{https://sites.google.com/view/reservoir-computing-tf/resources}} provide more details about these and other ESN toolboxes. Guang Bin Huang\footnote{\href{http://www.extreme-learning-machines.org/}{http://www.extreme-learning-machines.org/}} also developed a coherent summary of ELM implementations and there are many more individual projects accessible on GitHub\footnote{\href{https://github.com/topics/reservoir-computing}{https://github.com/topics/reservoir-computing}}.

A likely reason that RCNs are not yet officially included in leading Deep Learning toolkits, such as PyTorch or TensorFlow, is that their underlying training paradigm is very different from that of the conventional neural networks. Instead of using some kind of optimizer that requires a large number of iterations and a lot of parallel computing and electrical power, RCNs have only very few hyperparameters and are trained very efficiently using linear regression.

With this paper, we introduce PyRCN (Python Reservoir Computing Network), a new toolbox for RCNs that is based on widely used scientific Python packages, such as numpy or scipy, and complies with the interface specification of the popular machine learning platform scikit-learn \citep{src:Pedregosa-11}. 
As one original contribution of this work, this allows the seamless integration of PyRCN and scikit-learn and to benefit from the extensive list of scikit-learn features, such as model selection tools and alternatives to linear regression for training the output weights of RCNs. The designed models can also easily be compared with scikit-learns' built-in estimators, such as MLPs or Support Vector Machines (SVMs).

The second original contribution of this work is the definition of the \enquote{Building blocks of Reservoir Computing} as outlined in Section \ref{sec:BuildingBlocksOfReservoirComputing}. By splitting ELMs, ESNs and LSMs in these building blocks, almost any conceivable RCN architecture can be constructed. Since the internal structure of PyRCN is based on these building blocks, a large amount of RCN architectures can be constructed in only a few lines of code. This is a significant advantage of PyRCN over other active toolboxes for RCNs, which are specifically developed to provide only one specific type of RCN.

Another unique feature of PyRCN is the support for the hyper-parameter search strategy introduced in \citet{src:Steiner-20a}, which requires far fewer model fits than the conventional approach.

The remainder of this paper is structured as follows: In section \ref{sec:BuildingBlocksOfReservoirComputing}, we briefly review the main concepts of RCNs and show how to decompose them into basic building blocks. Section \ref{sec:PyRCN} introduces the proposed PyRCN toolbox with all included modules and shows a simple example of how to get started with PyRCN. In section \ref{sec:ReservoirComputingNetworks}, we show how to construct different RCNs from these components. Section \ref{sec:SequenceToProcessingWithESNs} demonstrates how to construct a custom RCN for a handwritten digit recognition task using PyRCN. Section \ref{sec:ComparisonOfToolboxes} compares PyRCN with the reference toolboxes PyESN and HP-ELM by re-implementing parts of the recent work by \citet{src:Trierweiler-Ribeiro-21}. Finally, we summarize our work and outline future work in section \ref{sec:ConclusionAndOutlook}.

\begin{landscape}
\begin{table}[!htb]
    \renewcommand{\arraystretch}{1.5}
    \centering
    \caption{Overview of important RCN toolboxes, download links, the year of the last major update and whether the toolboxes provide a high-level interface. The implementation of LSMs is currently under development (UD).}
    \label{tab:OverviewRCNToolboxesLandscape}
    \begin{tabular}{lllllll}
        \toprule
        \multirow{2}{*}{Toolbox} & \multirow{2}{*}{Online Source} & \multicolumn{3}{c}{Supported RCNs} & Last & High-level \\
         &  & ELM & ESN & LSM & updated & Interface \\
        \midrule
        ESNToolbox \citep{src:Jaeger-01a} & \citet{src:ESNToolbox} & -- & $\checkmark$ & -- & 2009 & -- \\
        DeepESN Toolbox \citep{src:Gallicchio-17} & \citet{src:DeepESNToolbox} & -- & $\checkmark$ & -- & 2019 & -- \\
        EchoTorch \citep{src:Schaetti-18} & \citet{src:EchoTorch} & -- & $\checkmark$ & -- & 2021 & -- \\
        PyTorch-ESN \citep{src:Gallicchio-17} & \citet{src:PyTorchESN} & -- & $\checkmark$ & -- & 2021 & -- \\
        ReservoirPy \citep{src:Trouvain-20} & \citet{src:ReservoirPy} & -- & $\checkmark$ & -- & 2022 & $\checkmark$ \\
        \midrule
        HP-ELM \citep{src:Akusok-15} & \citet{src:HpELM} & $\checkmark$ & -- & -- & 2018 & $\checkmark$ \\
        Python-ELM & \citet{src:Python-ELM} & $\checkmark$ & -- & -- & 2019 & $\checkmark$ \\
        \midrule
        LSM & \citet{src:LSM} & -- & -- & $\checkmark$ & 2020 & -- \\
        \midrule
        OGER \citep{src:Verstraeten-12} & \citet{src:Oger} & $\checkmark$ & $\checkmark$ & $\checkmark$ & 2013 & -- \\
        \midrule
        PyRCN (This work) & \citet{src:PyRCN} & $\checkmark$ & $\checkmark$ & UD & 2022 & $\checkmark$ \\
        \bottomrule
    \end{tabular}
\end{table}
\end{landscape}

\section{Building blocks of Reservoir Computing}
\label{sec:BuildingBlocksOfReservoirComputing}

A conventional Reservoir Computing Network (RCN) consists of various weight matrices that describe how input data is transported in and processed by a hidden layer, which is typically a pool of non-linear neurons. In order to provide a framework that implements existing architectures and enables desired customization, we need to design components that can be mixed and matched to create many different kinds of architectures instead of only offering static closed implementations of existing topologies. We postulate that all RC architectures can essentially be built by three main components (See Fig.\ \ref{fig:algOutline}): 

\begin{figure}[!htb]
	\centering
	\subfloat[ELM]{\includegraphics[height=.31\textwidth]{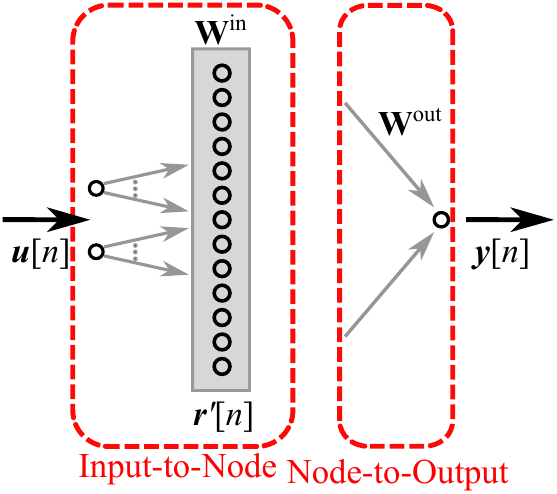}\label{subfig:ELM}}\ 
	\subfloat[ESN]{\includegraphics[height=.31\textwidth]{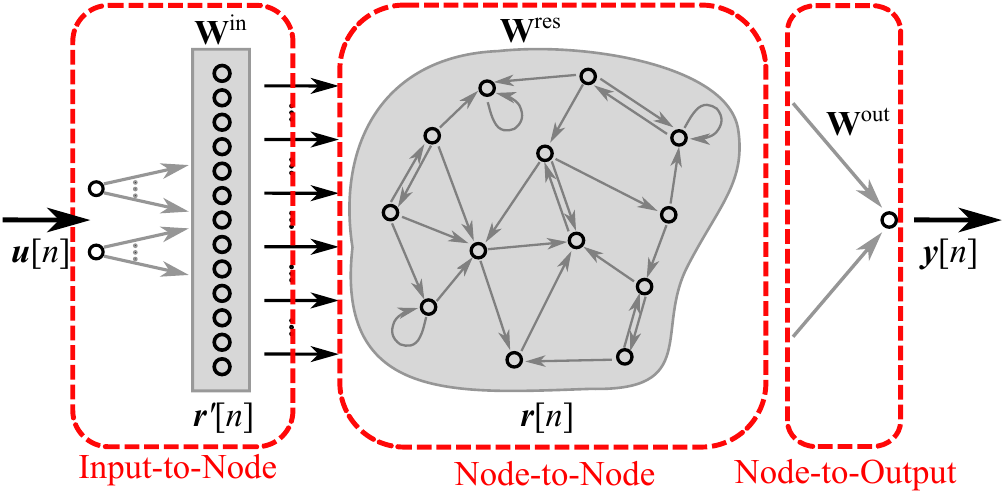}\label{subfig:ESN}}
	
	\subfloat[LSM]{\includegraphics[height=.31\textwidth]{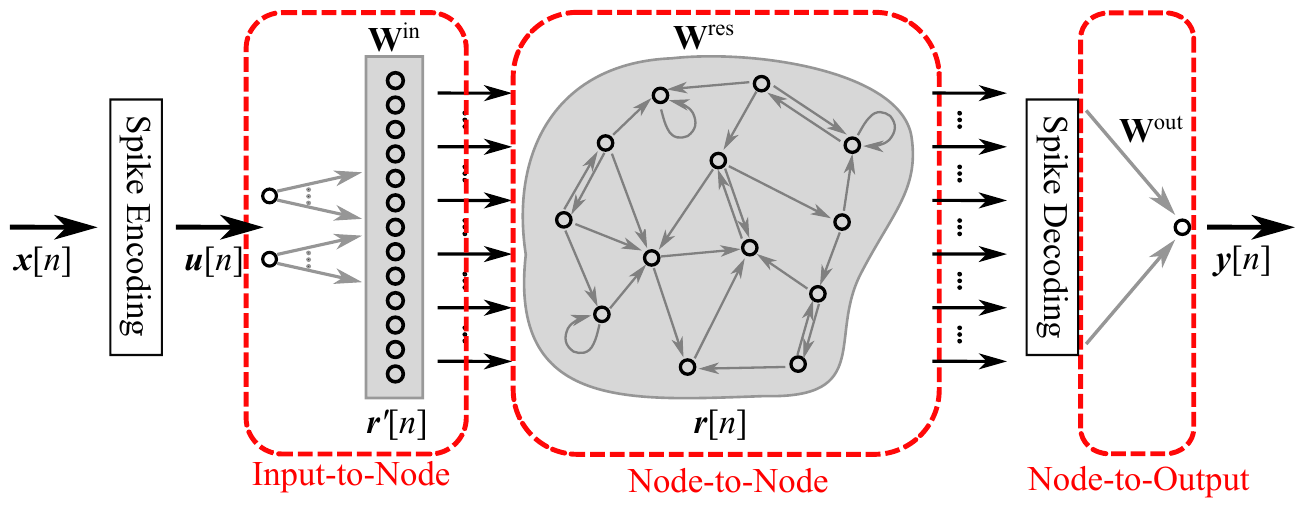}\label{subfig:LSM}}
	\caption{Main outline of basic RCNs: In case of ELMs and ESNs, the input features are fed into a hidden layer, which is called the \enquote{reservoir} using the fixed input weight matrix $\mathbf{W}^{\mathrm{in}}$. The reservoir consists of non-linear neurons. In case of the ESN, the neurons are inter-connected via the fixed reservoir matrix $\mathbf{W}^{\mathrm{res}}$. For both, ESNs and ELMs, the output $\mathbf{y}[n]$ is a weighted combination of the reservoir states $\mathbf{r}[n]$ based on the output weight matrix $\mathbf{W}^{\mathrm{out}}$. LSMs are closely related to ESNs. However, they need additional spike encoding and decoding utilities.}
	\label{fig:algOutline}
\end{figure}

\begin{enumerate}
    \item \enquote{Input-to-Node} that describes a combination of input weights and a non-linear input activation function, 
    \item \enquote{Node-To-Node} that describes recurrent reservoir weights and a non-linear reservoir activation function, and 
    \item \enquote{Node-to-Output} that describes the trainable output weights and is often referred to as linear model.
\end{enumerate}

Since LSMs use spiking neural networks, additional building blocks for spike encoding and decoding are required (see Fig.\ \ref{subfig:LSM})\footnote{For a simple notation, we assume that the sampling rate before and after spike en-/decoding is the same. In reality, the sampling rate after encoding is obviously much higher.} and are currently under development. However, the remaining blocks are essentially the same as for the ESN. 

Note that the black arrows in Fig.\ \ref{subfig:ESN} and \ref{subfig:LSM} between the building blocks all indicate one-by-one connections. E.g., the first black arrow of the ESN between \enquote{Input-to-Node} and \enquote{Node-to-Node} denotes that the first neuron in \enquote{Input-to-Node} is also the first neuron of \enquote{Node-to-Node}.

\subsection{Input-to-Node}
\label{subsec:InputToNode}

The \enquote{Input-to-Node} component describes the connections from the input features to the reservoir and the activation functions of the reservoir neurons. Normally, the input weight matrix $\mathbf{W}^{\mathrm{in}}$ has the dimension of $N^{\mathrm{res}}\times N^{\mathrm{in}}$, where $N^{\mathrm{res}}$ and $N^{\mathrm{in}}$ are the size of the reservoir and dimension of the input feature vector $\mathbf{u}[n]$ with the time index $n$, respectively. With 

\begin{align}
    \label{eq:InputToNode}
    \mathbf{r}'[n] = f'(\mathbf{W}^{\mathrm{in}}\mathbf{u}[n] + \mathbf{w}^{\mathrm{bi}}) \text{ , }
\end{align}
\noindent
we can describe the projection of the input features $\mathbf{u}[n]$ into the high-dimensional reservoir space $\mathbf{r}'[n]$ via the input activation function $f'(\cdot)$, which can be the identity function for the standard ESN (see Section \ref{subsec:EchoStateNetworks}) or any non-linear function for the ELM (see Section \ref{subsec:ExtremeLearningMachines}).

The values inside the input weight matrix are usually initialized randomly from a uniform distribution on the interval $[-1, 1]$ and are afterwards scaled using the input scaling factor $\alpha_{\mathrm{u}}$. Since in case of a high-dimensional input feature space and/or large reservoir sizes $N^{\mathrm{res}}$, this leads to a huge input weight matrix and expensive computations to feed the feature vectors into the reservoir, it was shown in \citet{src:Jaeger-01a,src:Jalalvand-15} that it is sufficient to have only a very small number of connections from the input nodes to the nodes inside the reservoir. Each node of the reservoir may therefore be connected to only $K^{\mathrm{in}}$ ($\ll N^{\mathrm{in}}$) randomly selected input entries. This makes $\mathbf{W}^{\mathrm{in}}$ typically very sparse and feeding the feature vectors into the reservoir potentially more efficient. The bias weights $\mathbf{w}^{\mathrm{bi}}$ with dimension $N^{\mathrm{res}}$ are typically initialized by fixed random values from a uniform distribution between $\pm 1$ and multiplied by the hyper-parameter $\alpha_{\mathrm{bi}}$.

\begin{lstlisting}[language=Python, caption={Minimal example for the \enquote{Input-to-Node} block.}, label=lst:InputToNode]
from pyrcn.base.blocks import InputToNode
from sklearn.datasets import make_blobs

# Generate a toy dataset
U, y = make_blobs(n_samples=100, n_features=10)
#      _ _ _ _ _ _ _ _
#     |               |
# ----| Input-to-Node |------
# u[n]|_ _ _ _ _ _ _ _|r'[n]
# U                    R_i2n

# Initialize, fit and apply an InputToNode
input_to_node = InputToNode(hidden_layer_size=50,
                            k_in=5, input_activation="tanh",
                            input_scaling=1.0, bias_scaling=0.1)

R_i2n = input_to_node.fit_transform(U)
print(U.shape, R_i2n.shape)
\end{lstlisting}

Listing \ref{lst:InputToNode} gives an example of how to use  \enquote{Input-to-Node}. The toy dataset with $N^{\mathrm{in}}=10$ is transformed in a $N^{\mathrm{res}}=50$-dimensional space via sparse input weights. Due to the input scaling factor $\alpha_{\mathrm{u}}=1$, the non-zero weights are uniformly distributed on the interval $[-1, 1]$. The non-zero bias scaling factor $\alpha_{\mathrm{bi}}$ leads to an additional constant bias input.

\subsection{Node-to-Node}
\label{subsec:NodeToNode}

The \enquote{Node-to-Node} component describes the connections inside the reservoir. The output of \enquote{Input-to-Node} $\mathbf{r}'[n]$ together with the output of \enquote{Node-to-Node} from the previous time step $\mathbf{r}[n-1]$ are used to compute the new output of \enquote{Node-to-Node} $\mathbf{r}[n]$ using 

\begin{align}
    \label{eq:NodeToNode}
    \mathbf{r}[n] = (1-\lambda)\mathbf{r}[n-1] + \lambda f(\mathbf{r}'[n] + \mathbf{W}^{\mathrm{res}}\mathbf{r}[n-1])\text{ , } 
\end{align}
\noindent
which is a leaky integration of the time-dependent reservoir states $\mathbf{r}[n]$. $f(\cdot)$ acts as the non-linear reservoir activation functions of the neurons in \enquote{Node-to-Node}. The leaky integration is equivalent to a first-order lowpass filter. Depending  on the leakage $\lambda \in (0, 1]$, the reservoir states are globally smoothed.

The reservoir weight matrix $\mathbf{W}^{\mathrm{res}}$ is a square matrix of the size $N^{\mathrm{res}}$. These weights are typically initialized from a standard normal distribution. The Echo State Property (ESP) requires that the states of all reservoir neurons need to decay in a finite time for a finite input pattern. In order to fulfill the ESP, the reservoir weight matrix is typically normalized by its largest absolute eigenvalue and rescaled to a spectral radius $\rho$, because it was shown in \citet{src:Jaeger-01a} that the ESP holds at least as long as $\rho \le 1$ but can also hold longer for specific hyper-parameter combinations \citep{src:Gallicchio-19}. The spectral radius and the leakage together shape the temporal memory of the reservoir. Similar as for \enquote{Input-to-Node}, the reservoir weight matrix gets huge in case of large reservoir sizes $N^{\mathrm{res}}$, it can be sufficient to only connect each node in the reservoir only to $K^{\mathrm{rec}}$ ($\ll N^{\mathrm{res}}$) randomly selected other nodes in the reservoir, and to set the remaining weights to zero.

Listing \ref{lst:NodeToNode} gives an example of how to use  \enquote{Node-to-Node}. As before, the toy dataset with $N^{\mathrm{in}}=10$ is transformed in a $N^{\mathrm{res}}=50$-dimensional space via \enquote{Input-to-Node}. Afterwards, the output is transformed in a dense \enquote{Node-to-Node} with recurrent connections (spectral radius $\rho=\num{1}$, leakage $\lambda=\num{0.9}$).

\begin{lstlisting}[language=Python, caption={Minimal example for the \enquote{Node-to-Node} block.}, label=lst:NodeToNode]
from pyrcn.base.blocks import NodeToNode
# Here goes the content of Listing 1
#      _ _ _ _ _ _ _ _        _ _ _ _ _ _ _
#     |               |      |              |
# ----| Input-to-Node |------| Node-to-Node |------
# u[n]|_ _ _ _ _ _ _ _|r'[n] |_ _ _ _ _ _ _ |r[n]
# U                    R_i2n                 R_n2n

# Initialize, fit and apply NodeToNode
node_to_node = NodeToNode(hidden_layer_size=50,
                          reservoir_activation="tanh",
                          spectral_radius=1.0, leakage=0.9,
                          bidirectional=False)
R_n2n = node_to_node.fit_transform(R_i2n)
print(U.shape, R_n2n.shape)
\end{lstlisting}

To incorporate information from the future inputs, bi-directional RCNs have been introduced, e.g.\ in \citet{src:Triefenbach-11}. The paradigm is realized when the output of \enquote{Node-to-Node} is computed in four steps:

\begin{enumerate}
    \item $\mathbf{r}_{\mathrm{fw}}[n]$ is computed using $\mathbf{r}'[n]$ forward with Eq.\ \eqref{eq:NodeToNode}.
    \item $\mathbf{r}_{\mathrm{bw}}[n]$ is computed using $\mathbf{r}'[n]$ reversed in time with Eq.\ \eqref{eq:NodeToNode}.
    \item Reverse $\mathbf{r}_{\mathrm{bw}}[n]$ again in time to obtain the original temporal order.
    \item Stack $\mathbf{r}_{\mathrm{fw}}[n]$ and $\mathbf{r}_{\mathrm{bw}}[n]$ to obtain the final reservoir state $\mathbf{r}[n]$ that includes future information.
\end{enumerate}

Utilizing \enquote{Node-to-Node} in the bi-directional mode would double the size of the reservoir state vector $\mathbf{r}[n]$ to $2 \times N_{\mathrm{res}}$.

\subsection{Node-to-Output}
\label{subsec:NodeToOutput}

The \enquote{Node-to-Output} component is the mapping of the reservoir state $\mathbf{r}[n]$ to the output $\mathbf{y}[n]$ of the network. In conventional RCNs, this mapping is trained using (regularized) linear regression. To that end, all reservoir states $\mathbf{r}[n]$ are concatenated into the reservoir state collection matrix $\mathbf{R}$. As linear regression usually contains an intercept term, every reservoir state $\mathbf{r}[n]$ is expanded by a constant of 1. All desired outputs $\mathbf{d}[n]$ are collected into the desired output collection matrix $\mathbf{D}$. Then, the mapping matrix $\mathbf{W}^{\mathrm{out}}$ can be computed using Eq.\ \eqref{eq:linearRegression}, where $\epsilon$ is a regularization parameter.

\begin{align}
    \label{eq:linearRegression}
    \mathbf{W}^{\mathrm{out}} =\left(\mathbf{R}\mathbf{R}^{\mathrm{T}} + \epsilon\mathbf{I}\right)^{-1}(\mathbf{D}\mathbf{R}^{\mathrm{T}})
\end{align}

The size of the output weight matrix $N^{\mathrm{out}}\times (N^{\mathrm{res}} + 1)$ or $N^{\mathrm{out}}\times (2 \times N^{\mathrm{res}} + 1)$ in case of a bidirectional \enquote{Node-to-Node} determines the total number of free parameters to be trained in the neural network. After training, the output $\mathbf{y}[n]$ can be computed using Equation \eqref{eq:readout}. 

\begin{align}
    \label{eq:readout}
    \mathbf{y}[n] = \mathbf{W}^{\mathrm{out}}\mathbf{r}[n]
\end{align}

Note that, in general, other training methodologies could be used to compute output weights. For example, in \citet{src:Triefenbach-11} it has been shown that a non-linear \enquote{Node-to-Output} can improve the performance of an RCN when limiting $N^{\mathrm{res}}$. Since solving Eq.\ \eqref{eq:linearRegression} is computationally expensive for large $N^{\mathrm{res}}$, PyRCN provides an incremental regression as proposed in \citet{src:Liang-06}.

\begin{lstlisting}[language=Python, caption={Minimal example for the \enquote{Node-to-Output} block.}, label=lst:NodeToOutput]
from sklearn.linear_model import Ridge
# Here goes the content of Listing 2
#       _ _ _ _ _ _ _       _ _ _ _ _ _ _        _ _ _ _ _ _ _        
#     |              |     |             |     |               |       
# ----|Input-to-Node |-----|Node-to-Node |-----|Node-to-Output |
# u[n]| _ _ _ _ _ _ _|r'[n]|_ _ _ _ _ _ _|r[n] | _ _ _ _ _ _ _ |
# U                   R_i2n               R_n2n        |
#                                                      |
#                                                 y[n] | y_pred

# Initialize, fit and apply NodeToOutput
y_pred = Ridge().fit(R_n2n, y).predict(R_n2n)
print(y_pred.shape)
\end{lstlisting}

Listing \ref{lst:NodeToOutput} gives an example of how to use  \enquote{Node-to-Output}. As before, the toy dataset with $N^{\mathrm{in}}=10$ is transformed in a $N^{\mathrm{res}}=50$-dimensional space via \enquote{Input-to-Node} and \enquote{Node-to-Node}. Finally, Eq.\ \eqref{eq:linearRegression} is solved using \lstinline{sklearn.linear_model.Ridge} with default settings.

\section{PyRCN}
\label{sec:PyRCN}

With PyRCN, we introduce a unified toolbox for implementing and developing different kinds of RCNs that is transparent and easy to use. PyRCN is built on scikit-learn and is fully compatible with its interface specification, allowing full interoperability between the toolkits. 

\subsection{Overview of PyRCN}
\label{subsec:OverviewOfPyRCN
}

\begin{figure}[!htb]
    \centering
    \resizebox{\textwidth}{!}{\input{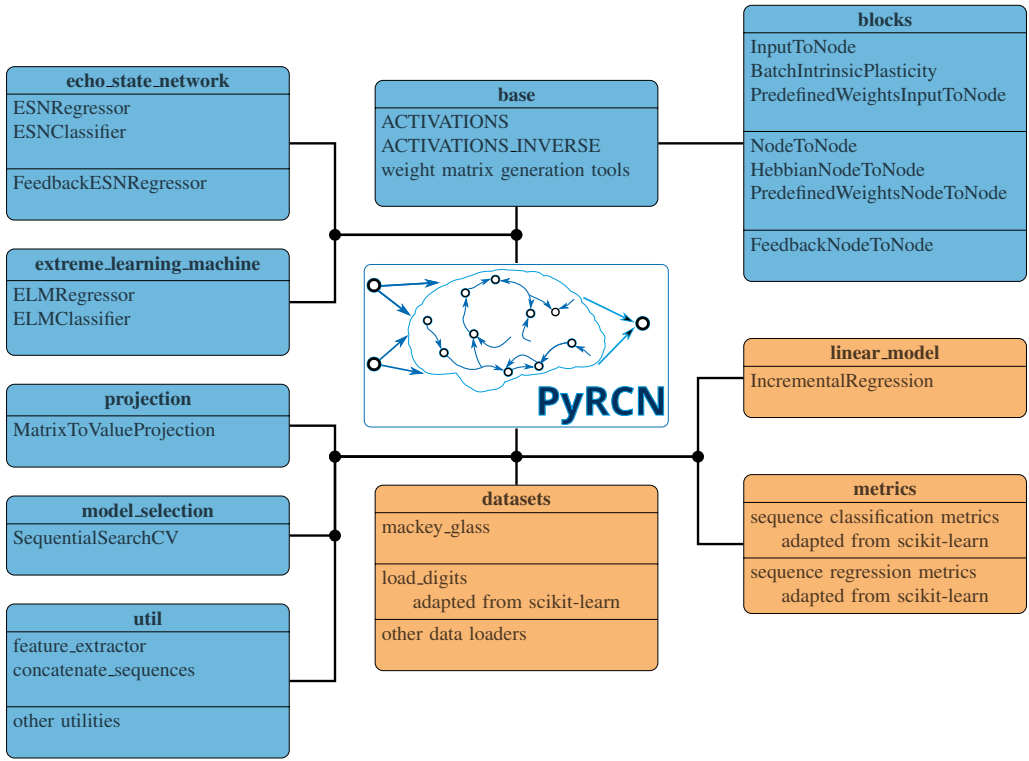}}
    \caption{Overview of the most important modules included in PyRCN. The blue modules include modules that are entirely new, whereas the orange modules can be treated as an extension of scikit-learn's built-in modules.}
    \label{fig:PyRCNOverview}
\end{figure}

The module \lstinline{pyrcn.base.blocks} provides classes that implement the two building blocks \enquote{Input-to-Node} (\lstinline{InputToNode}) and \enquote{Node-to-Node} (\lstinline{NodeToNode}) that are introduced in Section \ref{sec:BuildingBlocksOfReservoirComputing}. The module \lstinline{pyrcn.base} provides everything that is required to setting up own customized building blocks. These basic classes facilitate the random initialization of the two blocks. In addition, we provide trainable variants of both blocks, e.g.\ using BIP (\lstinline{BatchIntrinsicPlasticity}) or Hebbian Learning (\lstinline{HebbianNodeToNode}) \citep{src:Munakata-04}, and fully customizable variants (\lstinline{PredefinedWeights...}), where weight matrices can be set from outside. This offers the possibility to quickly define a standard RCN or to create very custom RCNs by defining own weight matrices such as in \citet{src:Rodan-11}.

As we will explain in Section \ref{sec:BuildingBlocksOfReservoirComputing}, the building block \enquote{Node-to-Output} is typically a linear regression. Thus, any regressor provided by scikit-learn can be used for this task. Since these cannot be trained incrementally, training on large datasets or training large networks becomes memory-expensive. To reduce the required memory, we provide a linear regression module based on \citet{src:Liang-06} in \lstinline{pyrcn.linear_model.IncrementalRegression}. Since the correlation matrices are updated incrementally, our linear regression module is capable of online learning, i.e., with the presentation of each new training sample.

In \lstinline{pyrcn.extreme_learning_machine} and \lstinline{pyrcn.echo_state_network}, we developed classes to build ELMs and ESNs for classification and regression following the scikit-learn API. In the different Listings of the Sections \ref{sec:BuildingBlocksOfReservoirComputing} and \ref{sec:ReservoirComputingNetworks}, we also show that these classes can be used to build a large variety of models.

To deal with sequence-to-value problems, we implemented the module \lstinline{pyrcn.projection} that allows to project from a matrix to values in different ways. This allows PyRCN to deal with sequential data while still staying as close as possible to the scikit-learn API (which does not natively support sequences).

In \lstinline{pyrcn.model_selection}, we developed the estimator \lstinline{SequentialSearchCV} as an extension for the existing model selection tools provided by scikit-learn. 

In \lstinline{pyrcn.metrics}, we have re-implemented most of the metrics from scikit-learn to handle sequence-to-sequence and sequence-to-label tasks. Using these metrics, all model selection techniques provided by scikit-learn can be applied to sequence learning tasks (not limited to using RCNs). 

Thanks to the scikit-learn interface, all modules can easily be customized and integrated. Furthermore, we steadily extend the library with new modules.

\subsection{Getting Started with PyRCN}
\label{subsec:GettingStartedWithPyRCN}

PyRCN can be installed with the command \lstinline{pip install pyrcn} or by cloning it from our Github repository\footnote{\href{https://github.com/TUD-STKS/PyRCN}{https://github.com/TUD-STKS/PyRCN}} and installing it with \lstinline{python setup.py install}. For more information, we refer to our website\footnote{\href{www.pyrcn.net}{www.pyrcn.net}}.

In Listing \ref{lst:minimumWorkingExample}, we show how to set up and train vanilla RCNs for predicting the Mackey-Glass time series \citep{src:Glass-10} with the same settings as used in \citet{src:Jaeger-01a} to introduce ESNs. The \lstinline{ESNRegressor} object connects the building blocks \enquote{Input-to-Node}, \enquote{Node-to-Node} and \enquote{Node-to-Output}, i.e., \lstinline{pyrcn.linear_model.IncrementalRegression}, and the \lstinline{ELMRegressor} object connects the building blocks \enquote{Input-to-Node} and \enquote{Node-to-Output}.
    
This minimum working example demonstrates two important aspects of PyRCN:
\begin{enumerate}
    \item Simplicity: it needs only four lines of code to load the Mackey-Glass dataset that is part of PyRCN and only two lines to fit the different RCN models, respectively.
    \item Interoperability with scikit-learn: Instead of the default incremental regression, we have customized the \lstinline{ELMRegressor} by using \lstinline{Ridge} from scikit-learn.
\end{enumerate}

\begin{lstlisting}[language=Python, caption={Minimal example for setting up an ESN and an ELM to predict the Mackey-Glass time-series \citep{src:Glass-10}}, label=lst:minimumWorkingExample]
from sklearn.linear_model import Ridge as skRidge
from sklearn.metrics import mean_squared_error
from pyrcn.echo_state_network import ESNRegressor
from pyrcn.extreme_learning_machine import ELMRegressor
from pyrcn.datasets import mackey_glass

# Load the dataset
X, y = mackey_glass(n_timesteps=5000)
# Define Train/Test lengths
trainLen = 1900
X_train, y_train = X[:trainLen], y[:trainLen]
X_test, y_test = X[trainLen:], y[trainLen:]

# Initialize and train an ELMRegressor and an ESNRegressor
esn = ESNRegressor().fit(X=X_train.reshape(-1, 1), y=y_train)
y_pred = esn.predict(X_test.reshape(-1, 1))
print(mean_squared_error(y_test, y_pred))
elm = ELMRegressor(regressor=skRidge()).fit(X=X_train.reshape(-1, 1), y=y_train)
y_pred = elm.predict(X_test.reshape(-1, 1))
print(mean_squared_error(y_test, y_pred))
\end{lstlisting}

\section{Build Reservoir Computing Networks with PyRCN}
\label{sec:ReservoirComputingNetworks}

By combining the building blocks introduced above, a vast number of different RCNs can be constructed. In this section, we build two important variants of RCNs, namely ELMs and ESNs.

\subsection{Extreme Learning Machines}
\label{subsec:ExtremeLearningMachines}

The vanilla ELM as a single-layer feedforward network consists of an \enquote{Input-to-Node} and a \enquote{Node-to-Output} module and is trained in two steps: 

\begin{enumerate}
    \item Compute the high-dimensional reservoir states $\mathbf{R}'$, which is the collection of reservoir states $\mathbf{r}'[n]$ from Equation \eqref{eq:InputToNode}.
    \item Compute the output weights $\mathbf{W}^{\mathrm{out}}$ using Equation \eqref{eq:linearRegression} with $\mathbf{R}'$.
\end{enumerate}

An example of how to construct such a standard ELM with PyRCN is given in Listing \ref{lst:BaseExtremeLearningMachine}. The \lstinline{ELMRegressor} internally passes the input features through \enquote{Input-to-Node} and trains \enquote{Node-to-Output} by solving Eq.\ \eqref{eq:linearRegression} via \lstinline{pyrcn.linear_model.IncrementalRegression}. 

\begin{lstlisting}[language=Python, caption={Example of how to construct a vanilla ELM with PyRCN.}, label=lst:BaseExtremeLearningMachine]
# Here goes the content of Listing 3
from pyrcn.extreme_learning_machine import ELMRegressor

# Vanilla ELM for regression tasks with input_scaling
#       _ _ _ _ _ _ _        _ _ _ _ _ _ _        
#     |              |     |               |       
# ----|Input-to-Node |-----|Node-to-Output |------
# u[n]| _ _ _ _ _ _ _|r'[n]| _ _ _ _ _ _ _ |y[n]
#                                           y_pred
# 
vanilla_elm = ELMRegressor(input_scaling=0.9)
vanilla_elm.fit(U, y)
print(vanilla_elm.predict(U))
\end{lstlisting}

When a custom \enquote{Input-to-Node} or \enquote{Node-to-Output} needs to be utilized, it is necessary to explicitly pass these to the ELM via \lstinline{input_to_node} or \lstinline{regressor}, respectively. This is, for example, necessary if the input weights should be pre-trained e.g.\ by Batch Intrinsic Plasticity (BIP) \citep{src:Neumann-11} or other unsupervised pre-training methods, such as in \citet{src:Steiner-21b}. An example for an ELM with a BIP \enquote{Input-to-Node} and a specific \enquote{Node-to-Output} is given in Listing \ref{lst:BIPExtremeLearningMachine}. Here, Eq.\ \eqref{eq:linearRegression} is solved using \lstinline{sklearn.linear_model.Ridge} instead of the built-in regression tool of PyRCN.

\begin{lstlisting}[language=Python, caption={Example of how to construct an ELM with a BIP \enquote{Input-to-Node} ELMs with PyRCN.}, label=lst:BIPExtremeLearningMachine]
# Here goes the content of Listing 3
from pyrcn.base.blocks import BatchIntrinsicPlasticity
from pyrcn.extreme_learning_machine import ELMRegressor

# Custom ELM with BatchIntrinsicPlasticity
#       _ _ _ _ _ _ _        _ _ _ _ _ _ _        
#     |              |     |               |       
# ----|     BIP      |-----|Node-to-Output |------
# u[n]| _ _ _ _ _ _ _|r'[n]| _ _ _ _ _ _ _ |y[n]
#                                           y_pred
# 
bip_elm = ELMRegressor(input_to_node=BatchIntrinsicPlasticity(),
                       regressor=Ridge(alpha=1e-5))

bip_elm.fit(U, y)
print(bip_elm.predict(U))
\end{lstlisting}

Hierarchical or Ensemble ELMs can then be built using multiple \enquote{Input-to-Node} modules in parallel or in a cascade. As can be seen in Listing \ref{lst:ComplexExtremeLearningMachines}, this is possible when using using scikit-learn's \lstinline{sklearn.pipeline.Pipeline} (cascading) or \lstinline{sklearn.pipeline.FeatureUnion} (ensemble). 

\begin{lstlisting}[language=Python, caption={Examples for implementing cascaded or Ensemle-ELMs with PyRCN.}, label=lst:ComplexExtremeLearningMachines]
# Here goes the content of Listing 3
from sklearn.pipeline import Pipeline, FeatureUnion
from pyrcn.extreme_learning_machine import ELMRegressor

# ELM with cascaded InputToNode and default regressor
#       _ _ _ _ _ _ _        _ _ _ _ _ _ _        _ _ _ _ _ _ _        
#     |     (bip)    |     |    (base)    |     |               |       
# ----|Input-to-Node1|-----|Input-to-Node2|-----|Node-to-Output |
# u[n]| _ _ _ _ _ _ _|     | _ _ _ _ _ _ _|r'[n]| _ _ _ _ _ _ _ |
#                                                       |
#                                                       |
#                                                  y[n] | y_pred
# 
i2n = Pipeline([('bip', BatchIntrinsicPlasticity()), 
                ('base', InputToNode(bias_scaling=0.1))])
casc_elm = ELMRegressor(input_to_node=i2n).fit(U, y)

# Ensemble of InputToNode with activations
#             _ _ _ _ _ _ _ 
#           |      (i)     |
#      |----|Input-to-Node1|-----|
#      |    | _ _ _ _ _ _ _|     |       _ _ _ _ _ _ _  
#      |                          -----|               |
# -----o                          r'[n]|Node-to-Output |------
# u[n] |      _ _ _ _ _ _ _      |-----| _ _ _ _ _ _ _ |y[n]   
#      |    |     (th)     |     |                      y_pred
#      |----|Input-to-Node2|-----|
#           | _ _ _ _ _ _ _|
# 
i2n = FeatureUnion([('i',InputToNode(input_activation="identity")), 
                    ('th',InputToNode(input_activation="tanh"))])
ens_elm = ELMRegressor(input_to_node=i2n)
ens_elm.fit(U, y)
print(casc_elm, ens_elm)
\end{lstlisting}

\subsection{Echo State Networks}
\label{subsec:EchoStateNetworks}

The Context Reverberation Network (CRN) \citep{src:Kirby-91} is a very early variant of Reservoir Computing that was later generalized to the ESN in \citet{src:Jaeger-01a}. CRNs and ESNs, as variants of RNNs, consequently consist of an \enquote{Input-to-Node}, a \enquote{Node-to-Node} and a \enquote{Node-to-Output} module and are trained in three steps. 

\begin{enumerate}
    \item Compute the neuron input states $\mathbf{R}'$, which is the collection of reservoir states $\mathbf{r}'[n]$ from Eq.\ \eqref{eq:InputToNode}. Note that here the activation function $f'(\cdot)$ is typically linear.
    \item Compute the reservoir states $\mathbf{R}$, which is the collection of reservoir states $\mathbf{r}[n]$ from Eq.\ \eqref{eq:NodeToNode}. Note that here the activation function $f(\cdot)$ is typically non-linear.
    \item Compute the output weights $\mathbf{W}^{\mathrm{out}}$ using
    \begin{enumerate}
        \item Eq.\ \eqref{eq:linearRegression} with $\mathbf{R}$ when considering an ESN.
        \item Gradient descent or other optimization algorithms when considering a CRN or when using an ESN with non-linear outputs, e.g.\ in \citet{src:Triefenbach-11}.
    \end{enumerate}
\end{enumerate}

An example of how to construct such a vanilla ESN with PyRCN is given in Listing \ref{lst:BaseEchoStateNetwork}, where the \lstinline{ESNRegressor} internally passes the input features through \enquote{Input-to-Node} and \enquote{Node-to-Node}, and trains \enquote{Node-to-Output} by solving Eq.\ \eqref{eq:linearRegression} via \lstinline{pyrcn.linear_model.IncrementalRegression}.

\begin{lstlisting}[language=Python, caption={Example of how to construct a vanilla ESN with PyRCN.}, label=lst:BaseEchoStateNetwork]
# Here goes the content of Listing 3
from pyrcn.echo_state_network import ESNRegressor

# Vanilla ESN for regression tasks with spectral_radius and leakage
#       _ _ _ _ _ _ _       _ _ _ _ _ _ _        _ _ _ _ _ _ _        
#     |              |     |             |     |               |       
# ----|Input-to-Node |-----|Node-to-Node |-----|Node-to-Output |
# u[n]| _ _ _ _ _ _ _|r'[n]|_ _ _ _ _ _ _|r[n] | _ _ _ _ _ _ _ |
#                                                      |
#                                                      |
#                                                 y[n] | y_pred
# 
vanilla_esn = ESNRegressor(spectral_radius=1, leakage=0.9)
vanilla_esn.fit(U, y)
print(vanilla_esn.predict(U))
\end{lstlisting}

As for ELMs in Listing \ref{lst:BIPExtremeLearningMachine},
various unsupervised learning techniques can be used to pre-train ESNs \citep{src:Basterrech-11,src:Schrauwen-08,src:Lazar-09}. For a summary, we refer to \citet{src:Steiner-21b}. An example is given in Listing \ref{lst:UnsupervisedESN}, in which we use a BIP \enquote{Input-to-Node} and a Hebbian \enquote{Node-to-Node}.

\begin{lstlisting}[language=Python, caption={Example of how to construct an ESN with a BIP \enquote{Input-to-Node} and a Hebbian \enquote{Node-to-Node} with PyRCN.}, label=lst:UnsupervisedESN]
# Here goes the content of Listing 3
from pyrcn.base.blocks import BatchIntrinsicPlasticity, HebbianNodeToNode
from pyrcn.echo_state_network import ESNRegressor

# Custom ESN with BatchIntrinsicPlasticity and HebbianNodeToNode
#       _ _ _ _ _ _ _       _ _ _ _ _ _ _        _ _ _ _ _ _ _        
#     |     (bip)    |     |   (hebb)    |     |               |       
# ----|Input-to-Node |-----|Node-to-Node |-----|Node-to-Output |
# u[n]| _ _ _ _ _ _ _|r'[n]|_ _ _ _ _ _ _|r[n] | _ _ _ _ _ _ _ |
#                                                      |
#                                                      |
#                                                 y[n] | y_pred
# 
bip_esn = ESNRegressor(input_to_node=BatchIntrinsicPlasticity(),
                       node_to_node=HebbianNodeToNode(),
                       regressor=Ridge(alpha=1e-5))

bip_esn.fit(U, y)
print(bip_esn.predict(U))
\end{lstlisting}

\subsection{Deep Echo State Networks}
\label{subsec:DeepEchoStateNetworks}

The term \enquote{Deep ESN} can refer to different approaches of hierarchical ESN architectures: 

\begin{itemize}
    \item The Deep ESN as described in \citet{src:Gallicchio-17} is a sequence of \enquote{Input-to-Node} and \enquote{Node-to-Node} combinations: the states of the $m-1$-th reservoir are the input for the $m$-th reservoir. The states of all reservoirs are stacked and based on the stacked states, the output weights are computed.
    \item The stacked ESN as used in \citet{src:Steiner-20b} is a sequence of \enquote{Input-to-Node}, \enquote{Node-to-Node} and \enquote{Output-to-Node} combinations: the outputs of one reservoir serve as the inputs for the next reservoir.
    \item The Modular Deep ESN as described in \citet{src:Carmichael-18} are multiple parallel \enquote{Input-to-Node} and \enquote{Node-to-Node} combinations: the output weights are computed over all parallel reservoir states.
\end{itemize}

With the proposed building blocks, we can build different variants of deep ESNs and many more customized variants. One rather complex example is given in Listing \ref{lst:DeepModularESN}, where we have a layer that mimicks the Modular Deep ESN in the first layer. On top of that, we have stacked a second reservoir that receives the output of the first layer as input.

\begin{lstlisting}[language=Python, caption={Example of how to construct a rather complex ESN consisting of two layers. It is built out of two small parallel reservoirs in the first layer and a large reservoir in the second layer.}, label=lst:DeepModularESN]
# Here goes the content of Listing 3
from pyrcn.base.blocks import InputToNode, NodeToNode
from pyrcn.echo_state_network import ESNRegressor

# Multilayer ESN
#                  u[n]
#                   |
#                   |
#          _________o_________
#         |                   |
#   _ _ _ | _ _ _       _ _ _ | _ _ _ 
# |      (i)     |    |      (i)     |
# |Input-to-Node1|    |Input-to-Node2|
# | _ _ _ _ _ _ _|    | _ _ _ _ _ _ _|
#         |r1'[n]             | r2'[n]
#   _ _ _ | _ _ _       _ _ _ | _ _ _
# |     (th)     |    |     (th)     |
# | Node-to-Node1|    | Node-to-Node2|
# | _ _ _ _ _ _ _|    | _ _ _ _ _ _ _|
#         |r1[n]              | r2[n]
#         |_____         _____|
#               |       |
#             _ | _ _ _ | _  
#           |               |
#           | Node-to-Node3 |
#           | _ _ _ _ _ _ _ |
#                   |
#              r3[n]|
#             _ _ _ | _ _ _  
#           |               |
#           |Node-to-Output |
#           | _ _ _ _ _ _ _ |
#                   |
#               y[n]|

l1 = Pipeline([('i2n1', InputToNode(hidden_layer_size=100)),
               ('n2n1', NodeToNode(hidden_layer_size=100))])

l2 = Pipeline([('i2n2', InputToNode(hidden_layer_size=400)),
               ('n2n2', NodeToNode(hidden_layer_size=400))])

i2n = FeatureUnion([('l1', l1),
                    ('l2', l2)])
n2n = NodeToNode(hidden_layer_size=500)
layered_esn = ESNRegressor(input_to_node=i2n,
                           node_to_node=n2n)

layered_esn.fit(U, y)
print(layered_esn.predict(U))
\end{lstlisting}

\subsection{Liquid State Machines}
\label{subsec:LiquidStateMachines}

Basic Liquid State Machines (LSMs) can be built in a similar way as ESNs. The main difference is that $\mathbf{u}[n]$ is a spike train sequence, hence, we need spiking neuronal models, and additional tools for spike coding and decoding as proposed in \citet{src:Schrauwen-03}, which are currently under development.
\subsection{Complex example: Optimize the hyper-parameters of RCNs}
\label{subsec:HyperparameterOptimization}

In \citet{src:Lukosevicius-12,src:Trouvain-20} it is discussed that the different hyper-parameters of RCNs are often tuned jointly using grid, random search or Bayesian Optimization \citep{src:Mockus-91}. However, this requires a lot of iterations, e.g., \num{1000} for the ESN model in \citet{src:Trouvain-20}. Other strategies, such as \citet{src:Jalalvand-15,src:Steiner-20a,src:Steiner-20b} proposed a half-guided sequential optimization procedure that can significantly reduce the number of iterations. 

While scikit-learn already provides utilities for performing grid and random searches, some effort is required to perform a sequential optimization. As can be seen in Listing \ref{lst:SequentialSearchCV}, we simplify this in PyRCN by introducing \lstinline{pyrcn.model_selection.SequentialSearchCV}, which is derived from the base search utilities of scikit-learn and is inspired by the \lstinline{sklearn.pipeline.Pipeline}. Essentially, a model with initial parameters is defined, followed by defining the different search steps. Similar to a \lstinline{sklearn.pipeline.Pipeline}, \lstinline{SequentialSearchCV} receives a list of searches (optimization steps), which are then sequentially performed.

Since internally, nothing else but the standard model selection tools provided by scikit-learn are evaluated, a particular advantage of \lstinline{SequentialSearchCV} is that it is not restricted to sequentially optimize PyRCN models but it can also be used to optimize any model that provides a scikit-learn interface. This shows how closely PyRCN and scikit-learn interact, and how it is possible to extend existing scikit-learn model selection tools. Note that it is furthermore possible to flexibly define an new optimization strategy, e.g., by changing the order of the optimization steps. 

\begin{lstlisting}[language=Python, caption={Example for a sequential parameter optimization with PyRCN. Therefore, a model with initial parameters and various search steps are defined. Internally, \lstinline{SequentialSearchCV} will perform the list of optimization steps sequentially.}, label=lst:SequentialSearchCV]
from sklearn.metrics import make_scorer
from sklearn.metrics import mean_squared_error
from sklearn.model_selection import TimeSeriesSplit
from sklearn.model_selection import RandomizedSearchCV, \
                                    GridSearchCV
from scipy.stats import uniform
from pyrcn.echo_state_network import ESNRegressor
from pyrcn.model_selection import SequentialSearchCV
from pyrcn.datasets import mackey_glass

# Load the dataset
X, y = mackey_glass(n_timesteps=5000)
X_train, X_test = X[:1900], X[1900:]
y_train, y_test = y[:1900], y[1900:]

# Define initial ESN model
esn = ESNRegressor(bias_scaling=0, spectral_radius=0, leakage=1,
                   requires_sequence=False)

# Define optimization workflow
scorer = make_scorer(mean_squared_error, greater_is_better=False)
step_1_params = {'input_scaling': uniform(loc=1e-2, scale=1),
                 'spectral_radius': uniform(loc=0, scale=2)}
kwargs_1 = {'n_iter': 200, 'n_jobs': -1, 'scoring': scorer, 
            'cv': TimeSeriesSplit()}
step_2_params = {'leakage': [0.2, 0.4, 0.7, 0.9, 1.0]}
kwargs_2 = {'verbose': 5, 'scoring': scorer, 'n_jobs': -1,
            'cv': TimeSeriesSplit()}

searches = [('step1', RandomizedSearchCV, step_1_params, kwargs_1),
            ('step2', GridSearchCV, step_2_params, kwargs_2)]

# Perform the search
esn_opti = SequentialSearchCV(esn, searches).fit(X_train.reshape(-1, 1), y_train)
print(esn_opti)
\end{lstlisting}

\section{Sequence processing with PyRCN}
\label{sec:SequenceToProcessingWithESNs}

ESNs are a variant of RNNs and, due to their recurrent connections, able to implicitly model temporal dependencies in the feature vector sequence, as it was discussed in Section \ref{subsec:EchoStateNetworks}. By default, scikit-learn does not support sequence handling. However, our incremental regression allows to deal with sequences of arbitrary lengths. With the following example, we show how to modify the programming pattern when dealing with sequences instead of instances.

\begin{lstlisting}[language=Python, caption={Sequence processing with PyRCN.}, label=lst:SequenceToLabel]
from sklearn.base import clone
from sklearn.model_selection import train_test_split
from sklearn.model_selection import ParameterGrid

from pyrcn.echo_state_network import ESNClassifier
from pyrcn.metrics import accuracy_score
from pyrcn.datasets import load_digits

# Load the dataset
X, y = load_digits(return_X_y=True, as_sequence=True)
print("Number of digits: {0}".format(len(X)))
print("Shape of digits {0}".format(X[0].shape))
# Divide the dataset into training and test subsets
X_tr, X_te, y_tr, y_te = train_test_split(X, y, test_size=0.2, 
                                          random_state=42)
print("Number of digits in training set: {0}".format(len(X_tr)))
print("Shape of the first digit: {0}".format(X_tr[0].shape))
print("Number of digits in test set: {0}".format(len(X_te)))
print("Shape of the first digit: {0}".format(X_te[0].shape))

# These parameters were optimized using SequentialSearchCV
esn_params = {'input_scaling': 0.1,
              'spectral_radius': 1.2,
              'input_activation': 'identity',
              'k_in': 5,
              'bias_scaling': 0.5,
              'reservoir_activation': 'tanh',
              'leakage': 0.1,
              'k_rec': 10,
              'alpha': 1e-5,
              'decision_strategy': "winner_takes_all"}

b_esn = ESNClassifier(**esn_params)

param_grid = {'hidden_layer_size': [50, 100, 200, 400, 500],
              'bidirectional': [False, True]}

for params in ParameterGrid(param_grid):
    esn = clone(b_esn).set_params(**params).fit(X_tr, y_tr)
    acc_score = accuracy_score(y_te, esn.predict(X_te))
\end{lstlisting}

\subsection{Handwritten Digits dataset}
\label{subsubsec:DigitsDataset}
For this task, we utilized the test set from the small handwritten digit dataset\footnote{\href{https://archive.ics.uci.edu/ml/datasets/Optical+Recognition+of+Handwritten+Digits}{https://archive.ics.uci.edu/ml/datasets/Optical+Recognition+of+Handwritten+Digits}} that is hosted by the UCI Machine Learning Repository \citep{src:Dua-17} and included in scikit-learn. It consists of \num{1797} \num{8}x\num{8} pixel gray-scale images of handwritten digits from 0 to 9. We normalized all pixels to fall in a range from \numrange{-1}{1}. For the following experiment, we treat each image of the dataset as an independent sequence that can, in general, have an arbitrary length. Therefore, we scan the image from left to right as in \citet{src:Jalalvand-15,src:Jalalvand-18} and thus present one column of eight pixels in one time step to the model.

\subsection*{Programming pattern for sequence processing}
\label{subsec:ProgrammingPatternForSequenceProcessing}

Listing \ref{lst:SequenceToLabel} shows the script for this experiment. This complex use-case requires a serious hyper-parameter tuning. To keep the code example simple, we did not include the optimization in this paper and refer the interested readers to the Jupyter Notebook\footnote{\href{https://github.com/TUD-STKS/PyRCN/blob/master/examples/digits.ipynb}{https://github.com/TUD-STKS/PyRCN/blob/master/examples/digits.ipynb}} that was developed to produce these results.

With \lstinline{pyrcn.datasets.load_digits}, we provide a wrapper function around the scikit-learn equivalent to load the digits dataset. Specifically, we added the option \lstinline{as_sequence}. If this option is set to \lstinline{True}, the dataset is returned as a numpy array of objects with a length of \num{1797} (number of images in the dataset). Each object itself is an $8\times 8$ numpy array representing the image as a sequence. This can be validated with the \lstinline{print} statements from line \numrange{11}{20} in Listing \ref{lst:SequenceToLabel}. 

Two particular advantages of this data structure are:

\begin{enumerate}
    \item We can use model selection utilities from scikit-learn to split our dataset in training and test sequences;
    \item Each sequence can have an arbitrary number of samples as long as the number of features is always the same.
\end{enumerate}

A disadvantage of this data structure is the break with scikit-learn's paradigm. However, since the number of features in every sequence is the same, we can simply reproduce e.g.\ the scikit-learn-compatible feature matrix via \lstinline{numpy.concatenate(X_train)}. This is important if one would compare ESNs with models provided by scikit-learn.

To ensure that the objects of PyRCN do not break with the programming pattern of scikit-learn when dealing with multiple sequences, we internally convert the new data structure in a scikit-learn compatible format and \lstinline{pyrcn.echo_state_network.ESNClassifier} can take the aforementioned data structure as well as the scikit-learn compatible data structure. Since scikit-learn does not provide any metrics for evaluating sequence processing tasks, we provide metrics, such as the accuracy score, for these tasks. These metrics can easily be combined with default model selection tools provided by scikit-learn. 

\subsection{Final results}
\label{subsec:FinalResults}

\begin{figure}[!htb]
    \centering
    \begin{tikzpicture}
\definecolor{clr1}{RGB}{195.8,181.7,56.1}
\definecolor{clr2}{RGB}{68.5,12.8,83.6}
\begin{axis}[
    width  = .9\textwidth,
    height = .3\textwidth,
    major x tick style = transparent,
    ymajorgrids = true,
    xmajorgrids = true,
    ylabel = {Accuracy},
    xlabel={$N_{\mathrm{res}}$},
    scaled x ticks = false,
    scaled y ticks = false,
    xmin=50,
    xmax=500,
    ymin=0.87,
    ymax=1.0,
    legend columns=2,
    legend style={at={(0.5,1.05)},anchor=south,font=\small},
]
\addplot[clr1, very thick, dashed, mark=x]
table {%
	50 0.91319444 0.89583333 0.8815331  0.89547038 0.90592334
	100 0.9375     0.92708333 0.93031359 0.95470383 0.90940767
	200 0.96527778 0.95138889 0.94425087 0.97212544 0.94076655
	400 0.95486111 0.97569444 0.96864111 0.9825784  0.95818815
	500 0.97222222 0.97569444 0.96515679 0.98606272 0.96515679
};\addlegendentry{ESN uni CV}
\addplot[clr1, very thick, mark=*]
table {%
	50 0.8916666666666667
	100 0.9277777777777778
	200 0.9638888888888889
	400 0.975
	500 0.9833333333333333
	1000 0.975
	2000 0.9527777777777777
};\addlegendentry{ESN uni Test}
\addplot[clr2, very thick, dashed, mark=x]
table {%
	50 0.87847222 0.90625    0.8815331  0.89198606 0.86759582
	100 0.96180556 0.95833333 0.92334495 0.96864111 0.93031359
	200 0.96527778 0.9375     0.95470383 0.98954704 0.94773519
	400 0.98263889 0.96527778 0.9825784  0.98606272 0.96167247
	500 0.97916667 0.96875    0.97212544 0.98606272 0.97560976
};\addlegendentry{ESN bi CV}
\addplot[clr2, very thick, mark=*]
table {%
	50 0.9055555555555556
	100 0.9444444444444444
	200 0.975
	400 0.9833333333333333
	500 0.9805555555555555
};\addlegendentry{ESN bi Test}
\end{axis}

\end{tikzpicture}
    \caption{Validation and test accuracy for the ESN model in uni- and bidirectional mode. The performance strongly increases until a reservoir size of \num{200} and remains almost constant afterwards.}
    \label{fig:ESN_MNIST_Evaluation}
\end{figure}
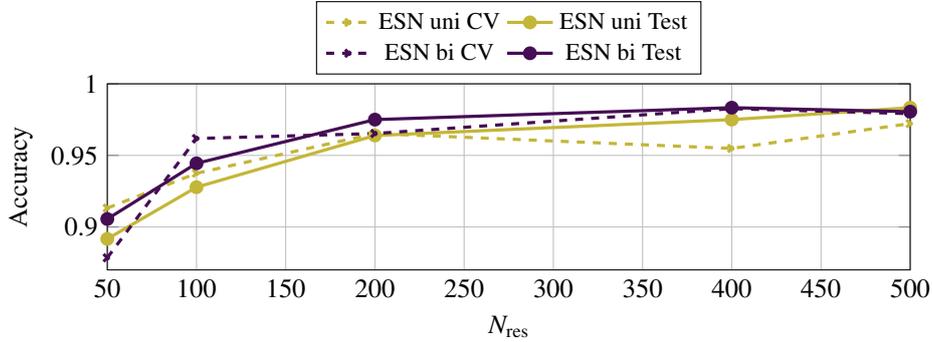

With the last few lines of code in Listing \ref{lst:SequenceToLabel}, we evaluate models with different reservoir sizes from \num{50} to \num{500} neurons and compared uni- and bidirectional ESNs. As demonstrated in \citet{src:Steiner-20a,src:Steiner-20b}, this is an important aspect of many experiments with RCNs, because these two steps significantly influence the final performance of RCN models. This is demonstrated in Figure \ref{fig:ESN_MNIST_Evaluation}, where the mean validation and the test accuracy strongly increases up to a reservoir size of \num{200} neurons and then slowly reaches saturation areas.

\section{Comparison of different toolboxes}
\label{sec:ComparisonOfToolboxes}

As it was stated in the introduction, RCNs have a broad variety of applications. As an example, we reproduce a part of the results by \citet{src:Trierweiler-Ribeiro-21}, where the stock price return volatility was predicted using Echo State Networks. We compare the performance of different RCN architectures included in PyRCN with the toolboxes PyESN\footnote{\href{https://github.com/cknd/pyESN}{https://github.com/cknd/pyESN}} and HP-ELM by \citet{src:Akusok-15} using the scikit-ELM\footnote{\href{https://github.com/akusok/scikit-elm}{https://github.com/akusok/scikit-elm}} (skELM) as the interface. All required data, pre-trained models and code to repeat the experiments can be found in our  repository\footnote{\href{https://github.com/TUD-STKS/PyRCN-Benchmark}{https://github.com/TUD-STKS/PyRCN-Benchmark}}.
    
\subsection{Dataset}
\label{subsec:DatasetStocPriceVolatility}

We used the datasets provided by \citet{src:Trierweiler-Ribeiro-21}, namely the stock price volatilities for the three Nasdaq companies Caterpillar (CAT), eBay (EBAY) and Microsoft (MSFT), with each datasets having a length of \num{2745} days. The input data was the original time series and the output data the time series shifted by 1, 5 and 22 days, respectively.

The time series were split in the same 3-fold cross validation scheme as in \citet{src:Trierweiler-Ribeiro-21} into training, validation and test sets according to Table \ref{tab:StockPriceSplit}. For the hyper-parameter optimization, only the training and validation sets of each fold were used. The test set of each fold was solely used to report the final results.

\begin{table}[!htb]
    \centering
    \caption{Partitioning the dataset in training, validation and test time series.}
    \label{tab:StockPriceSplit}
    \begin{tabular}{llll}
    \toprule
    Fold & Training & Validation & Test \\
    \midrule
    1 & CAT & EBAY & MSFT \\
    2 & EBAY & MSFT & CAT \\
    3 & MSFT & CAT & EBAY \\
    \bottomrule
    \end{tabular}
\end{table}

\subsection{Input and target preparation}
\label{subsec:InputAndTargetPreparation}

As in \citet{src:Trierweiler-Ribeiro-21}, we compared two different input data: The original time series and the Heterogeneous Autoregressive (HAR) feature set \citep{src:Corsi-09}, which expands the one-dimensional time series to three dimensions by adding moving-average filtered representations with filter lengths of five and 22 days, respectively. Since the description of the HAR feature set by \citet[Eqs.\ (4)--(7)]{src:Corsi-09} and \citet[Eq.\ (4)]{src:Trierweiler-Ribeiro-21} have different temporal contexts for the moving average filter, we compared both implementations. We use the description by \citet{src:Corsi-09} in the following, because the results were comparable to \citet{src:Trierweiler-Ribeiro-21}, even though not equal.
    
Each input and target time series was separately scaled to the interval $[0, 1]$ before using it to train and test models. The constants therefore were computed on the training dataset of each fold.
    
\subsection{Hyperparameter optimization}
\label{subsec:StockPriceHyperparameterOptimization}
    
In contrast to \citet{src:Trierweiler-Ribeiro-21}, we did not jointly optimize all hyper-parameters but used the sequential optimization from Section \ref{subsec:HyperparameterOptimization}. When considering ELMs, we aimed to optimize input scaling $\alpha_{\mathrm{u}}$, then bias scaling $\alpha_{\mathbf{bi}}$, and, finally, jointly the regularization parameter $\beta$ and the number of neurons $N^{\mathrm{res}}$. When considering ESNs, we aimed to jointly optimize input scaling $\alpha_{\mathrm{u}}$ and spectral radius $\rho$, then leakage $\lambda$, bias scaling $\alpha_{\mathrm{bi}}$, and, finally, jointly the regularization parameter $\beta$ and the number of neurons $N^{\mathrm{res}}$.
    
It is important to note that the different implementations did not always have the same hyper-parameters. The HP-ELM toolbox does neither provide a constant bias input nor support input scaling. Thus, we could not optimize the two parameters $\alpha_{\mathrm{u}}$ and $\alpha_{\mathrm{bi}}$, which we only optimized in PyRCN. The PyESN toolbox does not support leaky integration, and regularization, and does not provide a constant bias input. Thus, we could not optimize $\lambda$ and $\alpha_{\mathrm{bi}}$, which we only optimized in PyRCN. Instead of optimizing $\beta$ in PyESN, we optimized the noise level added to every neuron input. This is an alternative way to regularize ESNs \citep{src:Jaeger-01a}.
    
One advantage of PyRCN is its compliance with the scikit-learn interface specification. This becomes noticeable when we optimize the hyper-parameters with model selection tools provided by scikit-learn. Although the HP-ELM interface does not comply with scikit-learn, the scikit-elm\footnote{\href{https://github.com/akusok/scikit-elm}{https://github.com/akusok/scikit-elm}} adapter is available and we were able to use that for our experiments without any difficulties.
    
Such an adapter is not available for PyESN. We implemented it, which, however, slightly limits the power of PyESN, such as providing input scalings and input shifts with multiple dimensions (not required for this task c.f.\ \citet{src:Trierweiler-Ribeiro-21}). Implementing the adapter requires (a) knowledge about the scikit-learn interface and its underlying objects; (b) at least 64 unoptimized lines of code in accordance with the PEP-8 guidelines for Python. This is a disadvantage of PyESN (and other references in Table \ref{tab:OverviewRCNToolboxesLandscape}) compared to PyRCN. Alternatively, it would have been possible to implement an own randomized search routine, which, however, would require even more effort and would be less flexible than what \lstinline{sklearn.model_selection.RandomizedSearchCV} provides.
    
\subsection{Benchmark test}
\label{subsec:StockPriceBenchmark}

In the Tables \ref{tab:StockPriceH1}, \ref{tab:StockPriceH5} and \ref{tab:StockPriceH21}, the reference results from \citet{src:Trierweiler-Ribeiro-21} and the reproduced results by PyRCN and reference toolboxes are summarized. Overall, the results are similar. It is worth to note that \citet{src:Trierweiler-Ribeiro-21} used a modified variant of the PyESN toolbox for their experiments. However, in this paper, we used the original PyESN version and we used the same optimization scheme as for PyRCN. For all forecasting horizons, the results of PyESN and the ESN model in PyRCN are very similar. There is hardly any difference in the results between the ELM implementation of PyRCN and the HP-ELM. However, it can still be recognized that PyRCN outperformed the references in a majority of cases. This is mostly due to missing hyper-parameters such as leaky integration, a proper regularization or the bias scaling. PyRCN supports all these features and can thus be tuned for a task more thoroughly.

Interestingly, by limiting the optimization of the PyESN toolbox to the aforementioned hyper-parameters, we slightly outperformed \citet{src:Trierweiler-Ribeiro-21} in many cases. This suggests that our proposed way of sequential optimization from \citet{src:Steiner-20a,src:Steiner-21b} is also applicable for this kind of tasks.

\begin{table}[!htb]
    \centering
    \caption{Average values of volatility prediction metrics for 1-day ahead. The results of all models marked with a star symbol (*) are taken from \citep{src:Trierweiler-Ribeiro-21}, in which the PyESN toolbox was used. The differences between the literature and the results presented here arise from different parameter optimizations. Overall, the performance of different toolboxes was similar.}
    \label{tab:StockPriceH1}
    \begin{tabular}{lrrrrr}
        \toprule
        Model & $R^2_{\mathrm{train}}$ & $R^2_{\mathrm{test}}$ & $\mathrm{MSE}_{\mathrm{train}}$ & $\mathrm{MSE}_{\mathrm{test}}$ & $N^{\mathrm{res}}$ \\
         &  &  & \num{e-8} & \num{e-8} &  \\
        \midrule
        ESN\textsuperscript{*}\textsubscript{PyESN} & \num{0.634} & \num{0.632} & \num{5.74} & \num{5.81} & -- \\
        \midrule
        ESN\textsubscript{PyESN} & \num{0.646} & \num{0.619} & \num{5.81} & \num{5.92} & -- \\
        ESN\textsubscript{PyRCN} & \textbf{\num{0.681}} & \textbf{\num{0.644}} & \textbf{\num{5.23}} & \textbf{\num{5.62}} & \num{50} \\
        \midrule
        ELM\textsubscript{skELM} & \num{0.598} & \num{0.571} & \num{6.58} & \num{6.69} & \num{50} \\
        ELM\textsubscript{PyRCN} & \textbf{\num{0.609}} & \textbf{\num{0.578}} & \textbf{\num{6.41}} & \textbf{\num{6.62}} & \num{50} \\
        \midrule
        HAR\textsuperscript{*} & \num{0.650} & \num{0.633} & \num{5.78} & \num{5.81} & -- \\
        HAR &\num{0.653} & \num{0.636} & \num{5.69} & \num{5.73} & -- \\
        HAR-ESN\textsuperscript{*}\textsubscript{PyESN} & \num{0.637} & \num{0.635} & \num{5.75} & \num{5.78} & -- \\
        \midrule
        HAR-ESN\textsubscript{PyESN} & \num{0.657} & \textbf{\num{0.642}} & \num{5.58} & \num{5.64} & \num{50} \\
        HAR-ESN\textsubscript{PyRCN} & \textbf{\num{0.663}} & \num{0.639} & \textbf{\num{5.52}} & \num{5.67} & \num{50} \\
        \midrule
        HAR-ELM\textsubscript{skELM} & \num{0.654} & \num{0.636} & \num{5.68} & \num{5.72} & \num{400} \\
        HAR-ELM\textsubscript{PyRCN} & \textbf{\num{0.655}} & \num{0.636} & \textbf{\num{5.66}} & \num{5.72} & \num{50} \\
        \bottomrule
    \end{tabular}
\end{table}

\begin{table}[!htb]
    \centering
    \caption{Average values of volatility prediction metrics for 5-days ahead. The results of all models marked with a star symbol (*) are taken from \citep{src:Trierweiler-Ribeiro-21}, in which the PyESN toolbox was used. The differences between the literature and the results presented here arise from different parameter optimizations. Overall, the performance of different toolboxes was similar.}
    \label{tab:StockPriceH5}
    \begin{tabular}{lrrrrr}
        \toprule
        Model & $R^2_{\mathrm{train}}$ & $R^2_{\mathrm{test}}$ & $\mathrm{MSE}_{\mathrm{train}}$ & $\mathrm{MSE}_{\mathrm{test}}$ & $N^{\mathrm{res}}$ \\
         &  &  & \num{e-8} & \num{e-8} &  \\
        \midrule
        ESN\textsuperscript{*}\textsubscript{PyESN} & \num{0.485} & \num{0.480} & \num{8.27} & \num{5.81} & -- \\
        \midrule
        ESN\textsubscript{PyESN} & \num{0.500} & \num{0.476} & \num{8.25} & \num{8.35} & \num{50} \\
        ESN\textsubscript{PyRCN} & \textbf{\num{0.519}} & \textbf{\num{0.499}} & \textbf{\num{7.93}} & \textbf{\num{8.02}} & \num{100} \\
        \midrule
        ELM\textsubscript{skELM} & \num{0.432} & \num{0.406} & \num{9.37} & \num{9.55} & \num{50} \\
        ELM\textsubscript{PyRCN} & \textbf{\num{0.445}} & \textbf{\num{0.411}} & \textbf{\num{9.17}} & \textbf{\num{9.51}} & \num{50} \\
        \midrule
        HAR\textsuperscript{*} & \num{0.499} & \num{0.481} & \num{8.33} & \num{5.81} & -- \\
        HAR & \num{0.509} & \num{0.490} & \num{8.11} & \num{8.17} & -- \\
        HAR-ESN\textsuperscript{*}\textsubscript{PyESN} & \num{0.535} & \num{0.510} & \num{7.53} & \num{5.78} & -- \\
        \midrule
        HAR-ESN\textsubscript{PyESN} & \textbf{\num{0.557}} & \textbf{\num{0.513}} & \textbf{\num{7.29}} & \num{7.88} & \num{50} \\
        HAR-ESN\textsubscript{PyRCN} & \num{0.524} & \num{0.502} & \num{7.85} & \textbf{\num{7.96}} &\num{50} \\
        \midrule
        HAR-ELM\textsubscript{skELM} & \num{0.513} & \num{0.494} & \num{8.05} & \num{8.12} & \num{50} \\
        HAR-ELM\textsubscript{PyRCN} & \num{0.513} & \textbf{\num{0.495}} & \textbf{\num{8.04}} & \textbf{\num{8.11}} & \num{50} \\
        \bottomrule
    \end{tabular}
\end{table}

\begin{table}[!htb]
    \centering
    \caption{Average values of volatility prediction metrics for 21-days ahead. The results of all models marked with a star symbol (*) are taken from \citep{src:Trierweiler-Ribeiro-21}, in which the PyESN toolbox was used. The differences between the literature and the results presented here arise from different parameter optimizations. Overall, the performance of different toolboxes was similar.}
    \label{tab:StockPriceH21}
    \begin{tabular}{lrrrrr}
        \toprule
        Model & $R^2_{\mathrm{train}}$ & $R^2_{\mathrm{test}}$ & $\mathrm{MSE}_{\mathrm{train}}$ & $\mathrm{MSE}_{\mathrm{test}}$ & $N^{\mathrm{res}}$ \\
         &  &  & \num{e-7} & \num{e-7} &  \\
        \midrule
        ESN\textsuperscript{*}\textsubscript{PyESN} & \num{0.268} & \num{0.264} & \num{1.18} & \num{1.19} & -- \\
        \midrule
        ESN\textsubscript{PyESN} & \num{0.286} & \num{0.269} & \num{1.19} & \num{1.17} & \num{50} \\
        ESN\textsubscript{PyRCN} & \textbf{\num{0.362}} & \textbf{\num{0.316}} & \textbf{\num{1.06}} & \textbf{\num{1.10}} & \num{50} \\
        \midrule
        ELM\textsubscript{skELM} & \num{0.236} & \num{0.195} & \num{1.27} & \num{1.29} & \num{50} \\
        ELM\textsubscript{PyRCN} & \textbf{\num{0.256}} & \textbf{\num{0.218}} & \textbf{\num{1.24}} & \textbf{\num{1.26}} & \num{50} \\
        \midrule
        HAR\textsuperscript{*} & \num{0.256} & \num{0.224} & \num{1.25} & \num{1.27} & -- \\
        HAR & \num{0.295} & \num{0.261} & \num{1.17} & \num{1.19} & -- \\
        HAR-ESN\textsuperscript{*}\textsubscript{PyESN} & \num{0.297} & \num{0.298} & \num{1.11} & \num{1.16} & -- \\
        \midrule
        HAR-ESN\textsubscript{PyESN} & \num{0.263} & \num{0.229} & \num{1.23} & \num{1.25} & \num{50} \\
        HAR-ESN\textsubscript{PyRCN} & \textbf{\num{0.357}} & \textbf{\num{0.310}} & \textbf{\num{1.07}} & \textbf{\num{1.11}} & \num{50} \\
        \midrule
        HAR-ELM\textsubscript{skELM} & \num{0.300} & \num{0.268} & \num{1.17} & \num{1.18} & \num{50} \\
        HAR-ELM\textsubscript{PyRCN} & \textbf{\num{0.306}} & \textbf{\num{0.272}} & \textbf{\num{1.16}} & \textbf{\num{1.17}} & \num{50} \\
        \bottomrule
    \end{tabular}
\end{table}

\begin{figure}
    \centering
    \subfloat[ESN]{\includegraphics{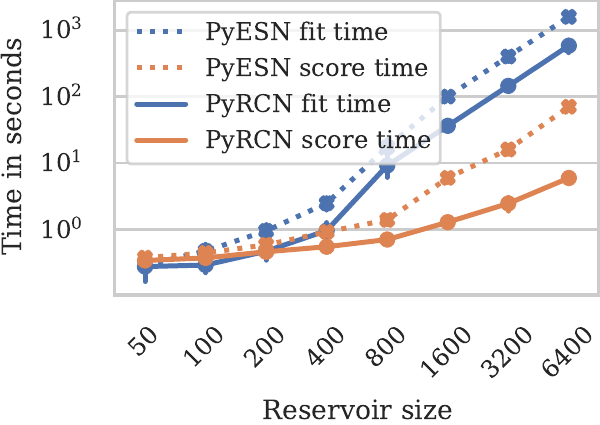}\label{subfig:ESNBenchmark}}
    \subfloat[ELM]{\includegraphics[trim=8 0 0 0, clip]{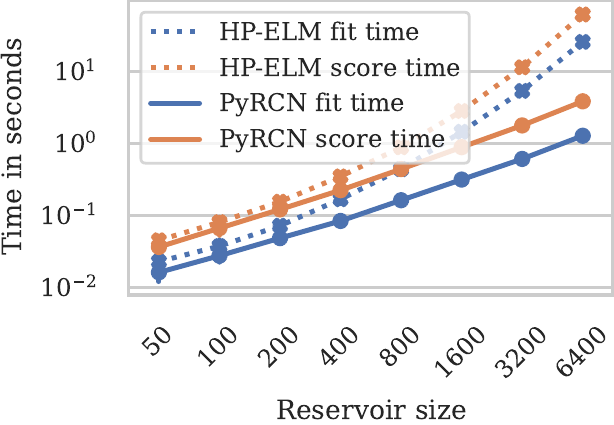}\label{subfig:ELMBenchmark}}
    \caption{Fit and score times for different RCN toolboxes. Especially for large ESNs, PyRCN requires less computational time than PyESN. Between PyRCN and HP-ELM, the difference gets smaller but is still present.}
    \label{fig:RCNBenchmark}
\end{figure}

Finally, we compare the fit and inference (score) times during the last step of the sequential hyper-parameter optimization, where we evaluated models with reservoir sizes between \num{50} and \num{6400} neurons and with different regularization methods. In Figure \ref{fig:RCNBenchmark}, the mean values and standard deviations of the fit and score times for different reservoir sizes are summarized for the different considered toolboxes. In all cases (ESNs and ELMs), the fit and score times of small models were similar. However, in case of large models, the differences between PyRCN and the reference toolboxes get larger and partially differ by more than one one order of magnitude, e.g.\ large ESNs (Fig.\ \ref{subfig:ESNBenchmark}). In fact, PyRCN is on average ten times faster than PyESN. One possible reason is that PyESN is typically trained with feedback connections back from the output to the reservoir. Since this requires extra computations in each time step, this is computationally more expensive, in particular for larger reservoirs. In case of the HP-ELM, the differences are lower. However, the required computational time of HP-ELM still increases with a larger slope than the one of PyRCN (Fig.\ \ref{subfig:ELMBenchmark}). Overall, this shows that PyRCN can be used for small and for large datasets and reservoirs.

\section{Conclusion and Outlook}
\label{sec:ConclusionAndOutlook}

We presented PyRCN, a new Python toolbox compatible with scikit-learn that offers flexible components to design many kinds of RCNs such as ESNs and ELMs. It is relatively light-weight with structured modules. It allows for convenient hyper-parameter optimization using standard scikit-learn model selection routines. An incremental regression as implemented in the toolkit performs equivalently as default ridge regression by scikit-learn while requiring less memory and decreasing the training time in case of large reservoirs. The toolkit supports both basic ESN models and sequence-to-sequence and sequence-to-label varieties and provides functionalities to utilize basically all built-in model selection techniques provided by scikit-learn also for sequence-to-sequence and sequence-to-label handling.

We have demonstrated how to use the toolbox on two widely known tasks, namely time-series prediction with the Mackey-Glass equation and handwritten digit classification using the handwritten digits dataset. Based on a benchmark test, we have shown that we can reproduce the results by \citet{src:Trierweiler-Ribeiro-21} using PyRCN with fewer lines of code, and that the sequential optimization scheme proposed and used by \citet{src:Steiner-20a,src:Steiner-20b,src:Steiner-21b} is applicable to univariate time-series. By comparing PyRCN with reference toolboxes, we have shown that the required computational time of PyRCN is comparable in case of small reservoirs and outperforms the reference toolboxes by decades in case of large reservoirs as PyRCN was in the mean ten times faster than PyESN.

A current limitation is that PyRCN does not provide an interface to other established machine learning frameworks, e.g., PyTorch or Keras. In the future, we will continue to work on further optimizations of the underlying methods of the toolbox, especially parallel sequence processing, multi-layer ESN functionalities, and consider exposing additional interfaces to support other established machine learning frameworks. We strongly encourage enthusiastic reservoir computing pracitioners and researchers to try out the toolbox and we welcome any feedback and contribution to improve it.

\section*{Acknowledgements}
Thanks to Michael Schindler for his valuable contribution to PyRCN during his diploma thesis about Extreme Learning Machines. Thanks to Gabriel Trierweiler-Robeiro for sharing the datasets used in \citep{src:Trierweiler-Ribeiro-21}.
 
This research was supported by Europäischer Sozialfonds (ESF), the Free State of Saxony (Application number: 100327771) and Ghent University under the Special Research Award number~BOF19/PDO/134.

\bibliography{refs}
\end{document}